\documentclass[10pt,twocolumn,letterpaper]{article}

\usepackage[pagenumbers]{cvpr} 

\definecolor{cvprblue}{rgb}{0.21,0.49,0.74}
\usepackage[pagebackref,breaklinks,colorlinks,allcolors=cvprblue]{hyperref}

\usepackage{algorithm}
\usepackage{algorithmic}

\usepackage{placeins}

\usepackage{multirow} 

\def\paperID{11545} 
\def\confName{CVPR}
\def\confYear{2025}

\title{CALA: A Class-Aware Logit Adapter for Few-Shot Class-Incremental Learning}

\author{
Chengyan Liu\textsuperscript{1\dag} 
\hspace{0.6cm} 
Linglan Zhao\textsuperscript{2\dag} \hspace{0.6cm} 
Fan Lyu\textsuperscript{3} \hspace{0.6cm} 
Kaile Du\textsuperscript{4} \hspace{0.6cm} 
Fuyuan Hu\textsuperscript{1*} \hspace{0.6cm} 
Tao Zhou\textsuperscript{6}
\\
\textsuperscript{1}School of electronic \& information engineering, Suzhou University of Science and Technology
\\
\textsuperscript{2}	Tencent Youtu Lab, \textsuperscript{3}	Chinese Academy of Sciences,
\textsuperscript{4}	Southeast University,
\textsuperscript{5}	North Minzu University
\\
{\tt\small chengyanliu@post.usts.edu.cn \hspace{0.6cm} fuyuanhu@mail.usts.edu.cn}
}

\begin{document}
\maketitle
\begin{abstract}
Few-Shot Class-Incremental Learning (FSCIL) defines a practical but challenging task where models are required to continuously learn novel concepts with only a few training samples. 
Due to data scarcity, existing FSCIL methods resort to training a backbone with abundant base data and then keeping it frozen afterward. 
However, the above operation often causes the backbone to overfit to base classes while overlooking the novel ones, leading to severe confusion between them. 
To address this issue, we propose Class-Aware Logit Adapter (CALA). 
Our method involves a lightweight adapter that learns to rectify biased predictions through a pseudo-incremental learning paradigm. 
In the real FSCIL process, we use the learned adapter to dynamically generate robust balancing factors. 
These factors can adjust confused novel instances back to their true label space based on their similarity to base classes. 
Specifically, when confusion is more likely to occur in novel instances that closely resemble base classes, greater rectification is required.
Notably, CALA operates on the classifier level, preserving the original feature space, thus it can be flexibly plugged into most of the existing FSCIL works for improved performance. 
Experiments on three benchmark datasets consistently validate the effectiveness and flexibility of CALA.
Codes will be available upon acceptance.
\end{abstract}    
\section{Introduction}

\renewcommand{\thefootnote}{\fnsymbol{footnote}}
\footnotetext[1]{: corresponding author. \textsuperscript{\dag}: authors contributed equally.}

In recent years, machine learning has achieved numerous impressive accomplishments in the computer vision field \cite{resnet, maskrcnn, swin, attention, mamba}. However, most of these accomplishments rely on large-scale datasets that have been pre-collected, such as Imagenet \cite{imagenet}. However, real-world scenarios often involve data arriving in continuous streams. Therefore \cite{CIL} proposes class-incremental learning, a learning paradigm \cite{replay1, replay2, reg1, reg2, para1, para2} that can learn new knowledge with the data stream arrives, and avoid forgetting old knowledge when learning new ones. Nevertheless, in real-world scenarios, it is often challenging to collect large amounts of labeled data. For example, in the medical field, the scarcity of certain diseases results in a limited amount of available labeled data. To overcome these limitations, Hong et al. \cite{TOPIC} propose the setting of Few-Shot Class-Incremental Learning (FSCIL), which is more practical given that data arrives in streams and only a few labeled examples are available.

\begin{figure}
\centering

\includegraphics[width=1\columnwidth]{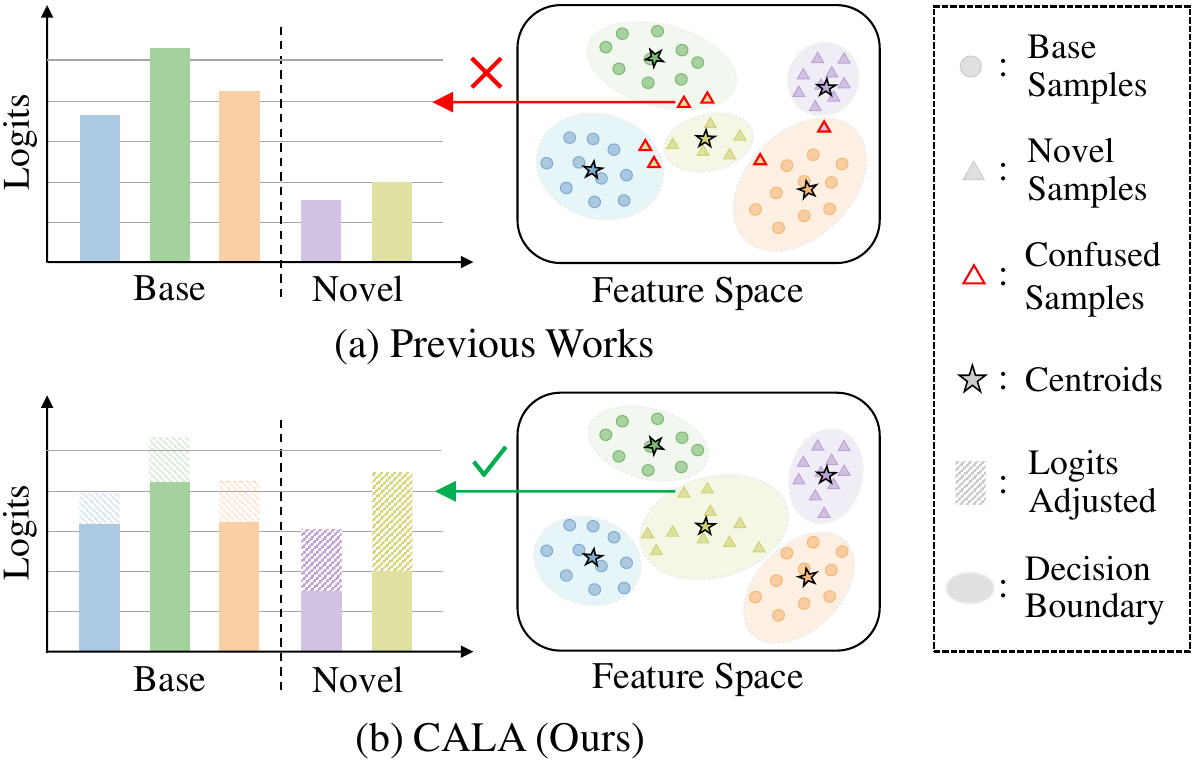} 
\caption{Comparisons of (a) previous works with the incremental-frozen framework and (b) our class-aware logit adaptor (CALA) for FSCIL from both the logit view and the feature space view.}
\label{fig1}
\end{figure}

Current prevalent FSCIL methods \cite{FACT, self, BiDist} mostly rely on an \textbf{incremental-frozen} framework \cite{savc}. 
In the above paradigm, a backbone is first trained on sufficient base class data and then frozen when learning novel classes with limited instances. 
Leading to the backbone over-fitting base classes while overlooking novel ones, \textit{i.e.}, the model shows a bias towards base classes. 
The bias persists throughout the entire classification process, leading to a distorted feature space, an imbalanced logit distribution, and ultimately erroneous prediction results. 
These issues favor the base classes while disadvantaging novel classes, as illustrated in Fig.~\ref{fig1} (a).
From the confusion matrix view, we call it \textbf{novel class confusion}, which leads to the model's performance in base classes overwhelmingly surpassing that in novel ones. Although the model shows ``good'' performance on base classes, this does not always ensure positive overall results, as the model is tested on all the ever-seen classes. 

The biased logit is a significant aspect contributing to confusion, which reminds us of post-hoc logit adjustment \cite{la}, a simple but effective strategy to balance logit distributions before classification \cite{longtail, L&GLA, vitlongtail}. 
We first propose a logit adjustment strategy for FSCIL by adding a class-agnostic factor $\mathbf{\alpha}$ to novel class logits, but it only shows limited improvements. 
It is because $\mathbf{\alpha}$ cursorily gives the same balance factors to all novel classes, while instances of different novel classes are influenced by base classes to varying degrees. 
As shown in Fig.~\ref{fig1}, since the yellow novel class is more distorted than the purple one, the former requires more logit balance weights than the latter. 
As a result, a class-agnostic factor can not meet the adjusting requirements for all the novel classes. 

Given that novel class confusion is a class-sensitive issue and cannot be resolved by class-agnostic factors alone, we build on the naive logit adjustment for FSCIL by proposing a \textbf{C}lass-\textbf{A}ware \textbf{L}ogit \textbf{A}daptor (CALA). 
CALA calculates a class-aware balance factor $\mathbf{\beta}$ using the similarity relationship between novel and base classes, as novel classes are more likely to be confused with similar base classes \cite{TEEN}. This approach allows us to dynamically correct the predictions of confused novel instances, as illustrated in Fig.~\ref{fig1} (b). Additionally, the greater the similarity between a novel class and base classes, the larger the correction applied by $\mathbf{\beta}$. 
To further improve its generalization, we envelop CALA in a pseudo-incremental learning paradigm to simulate the real FSCIL process, ensuring that CALA pre-acquires the ability to capture the similarity relationship between novel and base classes and calculate a more generalizable $\mathbf{\beta}$ with suppressed overfitting. 
Our contributions are threefold:

\begin{itemize}
\item To the best of our knowledge, we are among the first to observe the novel class confusion problem in FSCIL from the logit view and we propose a vanilla class-agnostic logit adjustment method for FSCIL to solve it.
\item Based on the above implementation, we further propose a \textbf{C}lass-\textbf{A}ware \textbf{L}ogit \textbf{A}dapter (CALA) module that dynamically corrects the logits of a novel class based on its similarity relationship with base classes, which can be used as a plug-and-play module applied to any incremental-frozen-based works, boosting the performance of traditional incremental-frozen framework.
\item Experiments on \textit{mini}-ImageNet, CIFAR-100, and CUB-200 show that our method can significantly outperform SOTAs. CALA can also be flexibly deployed to different FSCIL baselines to make considerable improvements.
\end{itemize}

\begin{figure*}
\centering
\includegraphics[width=1.9\columnwidth]{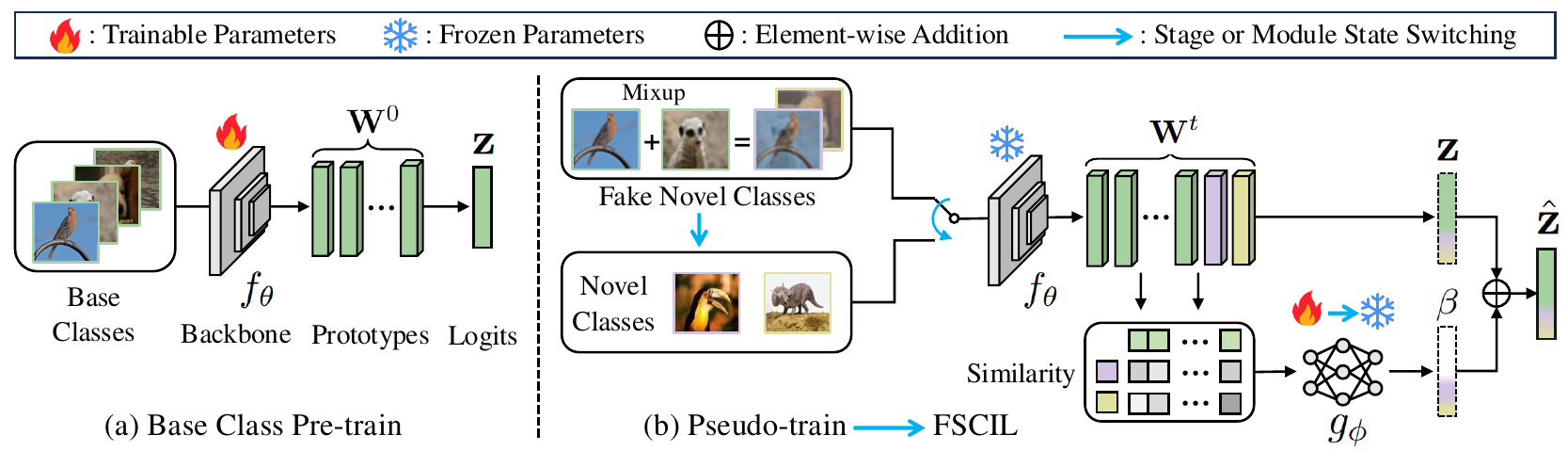} 
\caption{An overview of our method CALA. (a) In the base session, we use sufficient base data to pre-train a generalizable backbone that will be frozen in all subsequent processes. (b) In the upper branch, \textit{i.e.}, the pseudo-training stage, we mix up base data to create fake novel data and mimic an FSCIL process to train a relatively robust class-aware logit adapter. In the lower branch, \textit{i.e.}, during real FSCIL, we use the adapter to calculate a class-aware $\mathbf{\beta}$, and rectify the final logit in the testing stage.}
\label{figframework}
\end{figure*}
\section{Related Work}
\subsection{Few-Shot Learning}
Few-Shot Learning (FSL) addresses deep learning tasks with limited training data. Current FSL methods fall into three categories: Initialization-based approaches optimize pre-trained models to achieve a good starting point for FSL tasks \cite{maml, learning2learn}. Data augmentation-based methods use generative networks \cite{GAN} to create similar instances or embeddings \cite{featuretransfer, featurehallu}. Transfer learning-based ones leverage sufficient data to calibrate few-shot data distribution \cite{Freelunch, HOT} or introduce new metrics for classification \cite{protonet}. However, FSL typically overlooks sequentially arriving few-shot tasks.

\subsection{Few-Shot Class-Incremental Learning}
Few-Shot Class-Incremental Learning (FSCIL) aims to incrementally learn new knowledge from limited training instances, effectively making it a sequence of FSL tasks. This setting not only risks overfitting but also faces catastrophic forgetting. FSCIL was first introduced by TOPIC \cite{TOPIC}, which employed a neural gas network to learn new features while retaining old knowledge. Most popular methods \cite{metafscil, NCfscil, CEC} use meta-learning, simulating the FSCIL process with pseudo-incremental sessions to prepare the model in advance. Other approaches \cite{FACT, LIMIT, self, BiDist} adopt a \textbf{incremental-frozen} training framework, freezing most backbone parameters after training on base classes to prevent forgetting. However, TEEN \cite{TEEN} identified that this strategy causes the novel classes to be misclassified as base ones. TEEN addressed this by calibrating the novel feature distribution with the base feature distribution, though it may distort the original feature space. 
Notably, Our method CALA can suppress the confusion from the logit adjustment view and works on the classifier level, offering a plug-and-play solution that preserves the original features.

\subsection{Logit Adjustment}
The logit adjustment strategy, initially proposed in face recognition \cite{arcface, cosface}, has been effectively applied to long-tail learning \cite{GCL, TLA} due to its performance on imbalanced data. \cite{longtail} provides a comprehensive analysis and theoretical validation of this strategy. \cite{mclongtail} explores the link between logits and classification boundaries, balancing classification by optimizing logits. However, methods including \cite{L&GLA, vitlongtail} with a class-agnostic factor added to the logit term of loss function are not suitable for FSCIL, as the required balance weights for the logits of new classes differ from each other greatly. To the best of our knowledge, we are among the first to utilize logit adjustment for FSCIL tasks.
\section{Methodology}
\subsection{Problem Setting}
FSCIL typically involves $1 + T$ sessions, comprising one base session followed by $T$ incremental novel sessions, with data denoted as $D = \{D^0, D^1, \ldots, D^T\}$, where $D^0$ contains $B$ classes with abundant instances. For the $t$-th novel session ($t > 0$), the training data is denoted as $D^t = {(\mathbf{x}^t_i,\ y^t_i)}^{|D^t|}_{i=1}$, with $|D^t| \ll |D^0|$. $\mathbf{x}^t_i$ is the $i$-th instance in session $t$, $y^t_i$ is its corresponding label, where $y^t_i \in C^t$, the label space for session $t$. Each incremental session typically includes $N$ classes, with $K$ instances per class selected for training, known as $N$-way $K$-shot sampling where $|C^t|=N$ and $|D^t|=N K$. The label spaces across different sessions are disjoint, \textit{i.e.}, $\forall j, k \in [1, T]$ and $j \neq k, C^j \cap C^k = \varnothing$. Testing is conducted in each session, with all previously learned classes tested, \textit{i.e.}, the testing label space for session \(t\) is \(C^t_{\text{test}} = \bigcup_{j=0}^{t} C^j\).

\subsection{Overall Architecture}
Our main methods still follow the traditional incremental-frozen framework \cite{CEC, BiDist, savc} but enhance it by addressing the novel class confusion. The overall architecture of CALA is depicted in Fig.~\ref{figframework}. At first, a backbone is pre-trained on sufficient base data and then kept frozen in all subsequent processes, as shown in Fig.~\ref{figframework} (a). During the pseudo-training stage, fake novel classes are mixed up with base classes. In Section 3.3, we train a class-agnostic logit adjustment factor $\mathbf{\alpha}$ to balance all novel classes in the same weight in a relatively naive way. In Section 3.4, we improve Section 3.3 by training a class-aware logit adapter to calculate a corresponding balance factor $\mathbf{\beta}$ for each novel class considering their similarity relationship with base classes, as shown in the upper branch of Fig.~\ref{figframework} (b). When it comes to the real FSCIL stage, this adapter becomes frozen and used only to compute a class-aware logit adjustment factor $ \mathbf{\beta} $ based on the similarity matrix between novel and base classes, as shown in the lower branch of  Fig.~\ref{figframework} (b). Whether class-agnostic or class-aware, the logit adjustment factor is directly applied to the original logit during testing.

\subsection{Logit Adjustment for FSCIL}
Firstly, we generate fake novel classes with mixup \cite{mixup}, which generates fake novel instances by linearly interpolating pairs of base instances. For two instances $ \{(\mathbf{x}^0_i, \space y^0_i)\} $ and $ \{(\mathbf{x}^0_j, \space y^0_j)\} $, $y^0_i \neq y^0_j$, we mix them up to generate a fake novel instance $\mathbf{x}^{\tilde{t}}_k$ for pseudo-incremental session $ \tilde{t} $:

\begin{equation}
\mathbf{x}^{\tilde{t}}_k = \text{mixup}(\mathbf{x}^0_i, \space\mathbf{x}^0_j) = \lambda \cdot \mathbf{x}^0_i+(1-\lambda) \cdot \mathbf{x}^0_j \space,
\label{eq:1}
\end{equation}

\noindent $\lambda$ is a random number between 0.4 and 0.6. 
Each class pair is only used once to create $K$ fake novel instances for training and any number for testing from their instances, which repeats $N$ times for all fake novel classes in one session. Finally, pseudo-training data $D^{\tilde{t}} = {(\mathbf{x}_{k}^{\tilde{t}}, \space y_k^{\tilde{t}})}^{N K}_{k=1}$, $y^{\tilde{t}}_k = B + N  (\tilde{t}-1) + (k-1) // K$.

After the construction of  pseudo-training data, we compute the centroids fake novel classes in pseudo-incremental session $\tilde{t}$ with the feature embeddings $f_\theta(\mathbf{x}^{\tilde{t}}_i) \in \mathbb{R}^d$:

\begin{equation}
\mathbf{p}^{\tilde{t}}_c=\frac1{N_c}\sum_{i=1}^{|D^{\tilde{t}}|}\mathbb{I}(y^{\tilde{t}}_i=c)f_\theta(\mathbf{x}^{\tilde{t}}_i).
\label{eq:2}
\end{equation}

\noindent The vector $\mathbf{p}^{\tilde{t}}_c$, also known as the prototype of fake novel class $c$, which represents the most representative feature. $N_c$ denotes the number of instances in class $c$, $\mathbb{I}(\cdot)$ is the indicator function, $f_{\theta}(\cdot)$ represents the frozen backbone, which can effectively capture the semantic similarity between base and novel classes. The prototype $\mathbf{p}^{\tilde{t}}_c$ serves as the classifier for class $c$ in session $\tilde{t}$:

\begin{equation}
\mathbf{W}^{\tilde{t}} =[\mathbf{W}^{\tilde{t}-1}, \space{\{\mathbf{p}^{\tilde{t}}_c\}^N_{c=1}}],
\label{eq:3}
\end{equation}

\noindent $\mathbf{W}^0$ is a $d \times B$ weight matrix, initialized by concatenating the prototypes of all base classes. To update classifier $\mathbf{W} ^ {\tilde{t}}$, it is only necessary to concatenate $N$ prototypes of current session $\tilde{t}$ with classifier $\mathbf{W}^{\tilde{t}-1}$ from the last session.

Then we calculate original logits for $\mathbf{x} ^ {\tilde{t}}_i$ by:

\begin{equation}
{\mathbf{z}}=f_\theta(\mathbf{x}^{\tilde{t}}_i)\cdot\mathbf{W}^{\tilde{T}}.
\label{eq:4}
\end{equation}

\noindent Then we initialize a learnable parameter ${\alpha_c}$, we assume $\mathbf{\tilde{\alpha}} = {\alpha_c} \cdot \mathbf{1}$, where $\mathbf{1}$ is a $N\tilde{T}$-dimensional vector filled with 1, $\mathbf{\tilde{\alpha}}$ is the class-agnostic factor. We element-wisely add it to the novel class part in original logits $\mathbf{z}$. For each instance $\mathbf{x}_i$, ${\alpha_c}$ is optimized by the class-agnostic logit adjustment loss:

\begin{equation}
L_{LA}=-\sum_{j=1}^{N_C}\tilde{Y}_j\log\frac{e^{\mathbf{z}_j+[\mathbf{0},\space \tilde{\mathbf{\alpha}}]}}{\sum_{k=1}^{N_{c}}e^{\mathbf{z}_k+[\mathbf{0},\space \tilde{\mathbf{\alpha}}]}},
\label{eq:5}
\end{equation}
where $N_C$ is the number of classes, \textit{i.e.}, $N_C = B + N \tilde{T}$, $\tilde{Y}_j$ is the true label for $\mathbf{x_i}$ in class $j$.

The class-agnostic factor $\mathbf{\tilde{\alpha}}$ carries no class-aware information, it can only balance each novel class logit equally, which may cause under-balancing or over-balancing.

\subsection{Class-Aware Logit Adapter}
Considering the different requirements of logit balance between novel classes, under-balancing may still lead to a low performance on corresponding novel classes, meanwhile, over-balancing may transfer the bias on base classes to corresponding novel classes, confusing base instances into novel classes inversely.

Thus we propose a class-aware logit adapter to calculate a more robust factor after Eq.~\ref{eq:4}. In addition to serving as the classification weights for the base classes, $\mathbf{W}^0$ is also utilized to calculate the similarity matrix with the fake novel classes. We assume $\mathbf{W}^0 = [\mathbf{p}^0_1, \space \mathbf{p}^0_2, \space \dots, \space \mathbf{p}^0_B]$, then the similarity matrix $\tilde{\mathbf{S}}_c$ is defined as:

\begin{equation}
\tilde{\mathbf{S}}_c =[sim(\mathbf{p}^{\tilde{t}}_c,\space \mathbf{p}^0_1), \space sim(\mathbf{p}^{\tilde{t}}_c,\space \mathbf{p}^0_2), \space \dots,\space sim(\mathbf{p}^{\tilde{t}}_c,\space \mathbf{p}^0_B)].
\label{eq:6}
\end{equation}

\noindent We use cosine similarity as function $sim(\cdot)$. Since prototypes represent the most characteristic features of corresponding classes, \textit{i.e.}, the semantics, $sim(\mathbf{p}^{\tilde{t}}_c, \space \mathbf{p}^0_b)$ captures the semantic similarity between fake novel class $c$ and base class $b$. We then concatenate the semantic similarities of the novel class with all $B$ base classes to form the similarity matrix $\tilde{\mathbf{S}}_c$. A lightweight adapter takes this matrix as input, and its output is a scalar that serves as the balance factor element $\beta_c$ for the fake novel class $c$:

\begin{equation}
\beta_c=\tilde{g}_\phi(\sigma(\tilde{\mathbf{S}}_c)).
\label{eq:7}
\end{equation}

\noindent The adapter $\tilde{g}_\phi (\cdot)$ is a multi-layer perceptron (MLP) with two hidden layers, using the ReLU function as activation function. Since the fake novel class $c$ is constructed by mixing two base classes, it might exhibit high similarity to them. Therefore, before calculating $\tilde{\beta}$ with the adapter, the similarity matrix is first normalized using the softmax function $\sigma(\cdot)$. For each fake novel class in session $\tilde{t}$, a scalar $\beta_c$ is computed, and these scalars are concatenated into an $N$-dimensional vector $\mathbf{\beta}^{\tilde{t}}$. The $N$-dimensional vectors from each pseudo-incremental session are then concatenated in session order to form an $N\tilde{T}$ vector $\tilde{\mathbf{\beta}}$, representing the final balancing factors for all fake novel classes.

Due to the confusion among novel classes, the logits corresponding to base classes in ${\mathbf{z}}$ (the first $B$ elements) tend to be higher on average than those of the novel classes (the last $N \tilde{T}$ elements). To correct this imbalance, we add $\tilde{\mathbf{\beta}}$ element-wisely to the logits of the novel classes in ${\mathbf{z}}$:

\begin{equation}
\hat{\mathbf{z}}={\mathbf{z}}+[\mathbf{0},\space \gamma\tilde{\mathbf{\beta}}],
\label{eq:8}
\end{equation}
where $\mathbf{0}$ is a $B$-dimensional vector filled with $0$. The hyper-parameter $\gamma$ controls the extent of the adjustment by $\tilde{\mathbf{\beta}}$, helping to avoid both under-balancing and over-balancing. Since a novel instance is more likely to be confused into the base classes that are more similar to them,  $\tilde {\mathbf{\beta}}$ is computed based on the similarity matrix between fake novel and base classes, the greater the similarity between a fake novel class and all base classes, the larger adjustment $\tilde{\mathbf{\beta}}$ applies to that fake novel class.

After the logit adjustment of $\tilde{\mathbf{\beta}}$, the probability of fake novel instance $\mathbf{x}_i$ being predicted as class $j$ is: $\tilde{p}_j = \frac{e^{\mathbf{\hat{z}}_j}}{\sum_{k=1}^{N_C} e^{\mathbf{\hat{z}}_k}}$. The class-aware logit adapter loss function during the pseudo-training stage is:

\begin{equation}
L_{CALA}=-\sum_{j=1}^{N_C}\tilde{Y}_j\log(\tilde{p}_j)+\sum_{c=1}^{N  \tilde{T}}(\beta_c-\mu||\tilde{\mathbf{S}}_c ||_2).
\label{eq:9}
\end{equation}

After simulating the pseudo-training process, $\tilde{g}_\phi(\cdot)$ has learned the ability to capture the similarity relationship between fake novel and base classes and compute the appropriate class-aware balance factor $\mathbf{\beta}$, which can be transferred subsequently. Through pseudo-testing, we can obtain an adapter $g_\phi(\cdot)$ with relatively stronger generalization. In the real incremental learning stage, $g_\phi(\cdot)$ is frozen and used only to calculate the class-aware $\mathbf{\beta}$. 

The calculation of $\mathbf{\beta}$ during real incremental learning follows a process similar to the pseudo-incremental process. At first, prototypes are computed using Eq.~\ref{eq:2}. Next, the classifier $\mathbf{W}^{\tilde{t}}$ is updated using Eq.~\ref{eq:3}, and the similarity matrix $\mathbf{S}$ is calculated using Eq.~\ref{eq:6}. Finally, the adjusted logits $\hat{\mathbf{z}}$ are obtained using Eq.~\ref{eq:7} and Eq.~\ref{eq:8}.

During the testing stage\footnote{The utilization of $\tilde{\mathbf{\beta}}$ during the pseudo-testing phase is similar to that during the testing stage.}, the class-aware $\mathbf{\beta}$ calculated during the training process is also element-wisely added to the last $N T$ columns of the final logits, balancing the decision boundaries between novel and base classes and adjusting confused novel instances back to their own classes.

\textbf{Theory analysis.}
When using vanilla LA loss \cite{la} as the loss function, the Fisher consistency of the loss function is always supposed to be proved, \textit{i.e.}, in ideal conditions, the optimization of the objective function should be in line with the modification of the decision boundary. The general formula of vanilla LA loss is:

\begin{equation}
L_{VLA}=-\eta_j\log\left(\frac{e^{\mathbf{z}_j+\Delta_j}}{\sum_{k}e^{\mathbf{z}_k+\Delta_k}}\right).
\label{eq:10}
\end{equation}

\noindent We combine Eq.~\ref{eq:8} and $\tilde{p}_j = \frac{e^{\mathbf{\hat{z}}_j}}{\sum_{k=1}^{N_C} e^{\mathbf{\hat{z}}_k}}$, set $\eta_j = 1$ to get our own class-aware LA loss $L_{CALA}$. Through mathematical derivation, we simplify it to a more concise form for one single novel class:






\begin{equation}
    L_{CALA}=\log \left(1 + \sum_{k \neq j} e^{\mathbf{z}_k - \mathbf{z}_j}\right).
\label{eq:11}
\end{equation}

\noindent Assuming a novel class $j$, our method aims to enhance the logit of novel classes to make the decision boundary more favorable to novel classes, which corresponds with the increase of $\mathbf{z}_j$ and the decrease of the loss function, proving the Fisher consistency of our class-aware logit adapter loss.

\subsection{Pseudo Code and Discussions}

In the last section, we introduce the usage of CALA. Firstly, we pre-train a robust backbone $f_{\theta}(\cdot)$ on base data $D^0$, then we use it to calculate the prototypes of all base classes individually and construct the base weight $W^0$. We keep $f_{\theta}(\cdot)$ frozen in subsequent processes. 

\textbf{Pseudo Code.} We present the pseudo-training stage of CALA in Algorithm \ref{algcala}. For each pseudo-incremental session, we use mixup strategy for the construction of fake novel classes. Then we calculate the prototypes of every fake novel class and update the weight with them. We compute the similarity matrix between the base and novel weight, then we input it into the class-aware logit adapter $\tilde{g}_\phi (\cdot)$ and get the balance factor $\tilde{\mathbf {\beta}}$. We element-wisely add it to the logit and adjust the final results, by optimizing $L_{CALA}$ in Eq.~\ref{eq:9}. Finally, we get a new adapter ${g}_\phi (\cdot)$ with stronger generalizability. In the real FSCIL process, the application of ${g}_\phi (\cdot)$ and $\mathbf{\beta}$ are almost the same as that in the pseudo-training stage.

\begin{algorithm}[ht]
\caption{Pseudo-training of CALA}
\label{algcala}
\begin{algorithmic}[1]
\REQUIRE Pre-trained frozen backbone $f_{\theta}(\cdot)$, initialized adapter $\tilde{g}_\phi (\cdot)$, base data $D^0$, base weight $\mathbf{W}^0$.
\ENSURE A robust class-aware logit adapter ${g}_\phi (\cdot)$.
\WHILE{not done}
    \FOR{session=1, 2,\dots, $\tilde{T}$}
    \STATE Generate $N K$ fake novel instances by mixing up base instances with disjoint class pairs in Eq.~\ref{eq:1}.
    \STATE Calculate fake novel prototypes $\mathbf{p}^{\tilde{t}}_c$ in Eq.~\ref{eq:2} and update the classifier $\mathbf{W}^{\tilde{t}}$ with it in Eq.~\ref{eq:3}.
    \STATE Compute balance factor $\tilde{\mathbf{\beta}}$ with the cosine similarity between prototypes by $\tilde{g}_\phi (\cdot)$ in Eq.~\ref{eq:6}.
    \ENDFOR
    \STATE Calculate the logit with instance feature and $\mathbf{W}^{\tilde{t}}$, element-wisely add $\tilde{\mathbf{\beta}}$ to it in Eq.~\ref{eq:8}.
    \STATE 
Predict label, optimize $L_{CALA}$ to get ${g}_\phi (\cdot)$ in Eq.~\ref{eq:9}.
    
\ENDWHILE
\end{algorithmic}
\end{algorithm}

\textbf{Discussions.} TEEN \cite{TEEN} uses the distribution of base classes to calibrate the distribution of novel classes to suppress the novel class confusion, which is on feature level. However, CALA is on classifier level, which does no harm to the initial features, this is also why CALA can be considered as a plug-and-play module and employed on other works based on the incremental-frozen framework.
\section{Experiments}

In this section, we first introduce the specific experiment setting of FSCIL. Then we compare CALA on three prevailing benchmarks with State-of-The-Arts (SOTAs). Besides, ablations and plug-and-play experiments are conducted to show the advantages of CALA. 

\subsection{Datasets}
We follow the traditional setting of \cite{TOPIC} and conduct experiments on 3 different prevailing benchmarks:
\begin{itemize}
    \item \textbf{\textit{mini}-ImageNet} \cite{miniimagenet} is a subset of ImageNet \cite{miniimagenet}, with 100 classes and 600 84$\times$84 RGB images for each class. We split the 100 classes into 60 base classes and 40 novel classes just like \cite{TOPIC}.
\end{itemize}

\begin{itemize}
    \item \textbf{CIFAR100} \cite{cifar100} includes 100 classes with 60,000 tiny images of size 32$\times$32. The split of CIFAR 100 is just as same as that of \textit{mini}-ImageNet, \textit{i.e.}, 60 base classes and eight 5-way 5-shot classification incremental sessions, summing up to 40 novel classes.
\end{itemize}

\begin{itemize}
    \item \textbf{CUB200} \cite{cub200}, also known as the Caltech-UCSD Birds-200 dataset, contains 200 bird species, each with detailed, fine-grained characteristics. It owns 11,788 images, each image has a resolution of 224×224. Following \cite{TOPIC} we separate the 200 classes into 100 base classes and 10 novel incremental sessions, each of them has a 10-way 5-shot classification task.
\end{itemize}

\subsection{Implementation Details}
We conduct our experiments with the Pytorch library, we use ResNet18 as the backbone for the above three datasets, and we use SGD with momentum for optimization. In the base session, we follow \cite{BiDist} to pre-train the backbone. Specifically, we conduct a 200-epoch training with an initial learning rate of 0.1 for \textit{mini}-ImageNet and CIFAR100, the milestones are 120 and 160, and the decay rate is set as 0.1. The CUB200 model starts with ResNet18 pre-trained with ImageNet and undergoes 120 additional epochs of training with an initial learning rate of 0.01, the milestones are 50, 70, and 90, and the decay rate is set as 0.1, the normalization is specified for CUB200. Resizing, random crop, and random horizontal flip are used for data augmentation. During the pseudo-training stage, we allocate 64 hidden units to each layer of the two-layer MLP, the learning rate of MLP is 0.001, and we use fake novel data, constructed by base data, to mimic incremental sessions through 20 epochs.

\begin{table}[h]
\scriptsize
\centering
\renewcommand{\arraystretch}{1.25} 
\setlength{\tabcolsep}{2pt} 
\resizebox{1\linewidth}{!}{
 \begin{tabular}{l cccccccc }
  \toprule
  \multirow{2}{*}{Methods} &\multicolumn{8}{c}{FPR in each session (\%)}\\
  \cmidrule{2-9} 
   &{1}  &{2} &{3} &{4} &{5}  &{6} &{7} &{8}\\ 
  \midrule             
  
CEC~\cite{CEC}   
 &68.34 &65.49 &61.90 &58.26 &56.65 &54.46 &52.51 &50.02 \\

FACT~\cite{FACT} 
 &72.35 &70.54 &70.09 &69.36 &68.24 &68.13 &66.48 &66.51 \\

BiDist~\cite{BiDist}
  &78.20 &75.00 &72.46 &70.90 &69.12 &67.80 &64.97 &64.12 \\
  \midrule

Baseline  
  &72.60 &68.40 &65.60 &64.85 &61.40 &59.57 &56.03 &54.52 \\

LA-fscil (Ours)
  &65.40 &58.32 &55.24 &53.92 &52.33 &50.51 &47.08 &44.79 \\
{CALA (Ours)} 
&\textbf{52.60} &\textbf{42.10} &\textbf{40.60} &\textbf{40.30} &\textbf{35.92} &\textbf{33.57} &\textbf{29.57} &\textbf{28.40} \\
  
\bottomrule
\end{tabular}}
\caption{Ablations of our method using False Positive Rate (lower is better) on \textit{mini}-imageNet. We start the analysis from session 1 because there are no novel classes in session 0. Higher FPR means novel instances are more likely misclassified into base classes. CALA achieves a lower FPR than other methods and our own baseline, validating the effectiveness of our method.
}
\label{tabminiFNR}
\end{table}

\begin{table*}[ht]
\scriptsize
\centering
\renewcommand{\arraystretch}{1.2} 
\resizebox{0.93\linewidth}{!}{
 \begin{tabular}{l cccccccccccc cc  }
  \toprule
  \multirow{2}{*}{Method} &\multicolumn{9}{c}{Accuracy in each session(\%)} & \multirow{2}{*}{Avg} & \multirow{2}{*}{$\Delta_{\text{last}}$}\\
  \cmidrule{2-10}
  & {0} &{1}  &{2} &{3} &{4} &{5}  &{6} &{7} &{8}& & \\ 
  \hline              

  iCaRL~\cite{icarl}              &61.31 & 46.32& 42.94& 37.63& 30.49& 24.00& 20.89& 18.80& 17.21& 33.29 &  \textbf{+40.54}  \\
  
  TOPIC~\cite{TOPIC}              &61.31 &50.09& 45.17& 41.16& 37.48& 35.52& 32.19& 29.46& 24.42 &39.64 &  \textbf{+33.33}  \\
  
    CEC{$^\diamond$}\cite{CEC}     & 72.17& 66.97 & 62.96 &59.50 &56.62 &53.78 &51.15 &49.32 &47.45 &57.77 &  \textbf{+10.30} \\
  
  FACT{$^\diamond$}\cite{FACT}    & 72.71& 69.78& 66.53& 62.81& 60.73& 57.46& 54.46& 52.28& 50.60  & 60.82 & \textbf{+7.15} \\
  
  LIMIT~\cite{LIMIT}                &  72.32 &68.47 &64.30 &60.78 &57.95 &55.07 &52.70 &50.72 &49.19&  59.05 &\textbf{+8.56}  \\
    
  ALICE~\cite{ALICE} &80.60 &70.60 &67.40 &64.50 &62.50 &60.00 &57.80 &56.80 &55.70  &63.99   & \textbf{+2.05}     \\
    
  BiDist{$^\diamond$}\cite{BiDist}  &74.60 &69.85 &65.30  &61.67 &58.65 &55.47 &52.77 &50.82 &49.01 &59.79  & \textbf{+8.74}\\
  
  GKEAL~\cite{gkeal}  &73.59  &68.90  &65.33  &62.29  &59.39  &56.70  &54.20  &52.59  &51.31  &60.47  &\textbf{+6.44}    \\

  SAVC{$^\diamond$}~\cite{savc}  &80.47  &75.55  &71.26  &67.66  &64.49  &60.50  &57.90  &55.75  &54.04  &65.29  & \textbf{+3.71}\\
  
  TEEN~\cite{TEEN}   &73.53  &70.55  &66.37  &63.23  &60.53  &57.95  &55.24  &53.44  &52.08  &61.44   & \textbf{+5.67}     \\

  \midrule
 {LA-fscil (Ours)}   &83.98  &78.60  &73.95  &69.22  &65.85  &62.95  &59.68  &58.11  &56.65  &65.43  & \textbf{+1.10}     \\
  
  {CALA (Ours)}   &  \textbf{83.98}  &  \textbf{79.09}  &  \textbf{74.70}  &  \textbf{70.92}  &  \textbf{67.46}  &  \textbf{64.25}  &  \textbf{61.26}  &  \textbf{59.32}  &  \textbf{57.75} & \textbf{68.75}\\
  \bottomrule
\end{tabular}}
\caption{Comparison with the state-of-the-art of each session on the \textit{mini}-ImageNet dataset. Results with $\diamond$ are reproduced based on their open-source codes, and the other results are from their own articles. $\Delta_{\text{last}}$ means the improvement in the last session. Please refer to our supplementary material for more detailed results on the other two datasets.}
\label{tabminicom}	
\end{table*}

\subsection{Comparisons with SOTAs}
We conduct a toy experiment on \textit{mini}-ImageNet following \cite{TEEN} to show the ability of CALA to adjust novel instances confused as base classes back to novel ones. We transform the FSCIL into a binary classification problem. Specifically, we consider all base classes as positive classes while all novel classes as negative classes. Then we calculate the false positive rate (FPR) in this binary classification setting: $\text{FPR}=\frac{\text{FP}}{\text{FP}+\text{TN}}\times100\%$, which represents the probability that novel instances are confused as base classes.

As shown in Table~\ref{tabminiFNR}, our methods have lower FPR than other baselines in each session, namely, they are less likely to confuse novel instances into base classes. Although our class-agnostic logit adjustment for FSCIL (LA-fscil) achieves a lower FPR than other baselines, it still surpasses CALA, which means a class-agnostic factor cannot balance the confused novel instances back to their own classes effectively. In conclusion, CALA fully aligns with its motivation and is relatively more effective.

\begin{figure}[h]
\centering
\includegraphics[width=1\columnwidth]{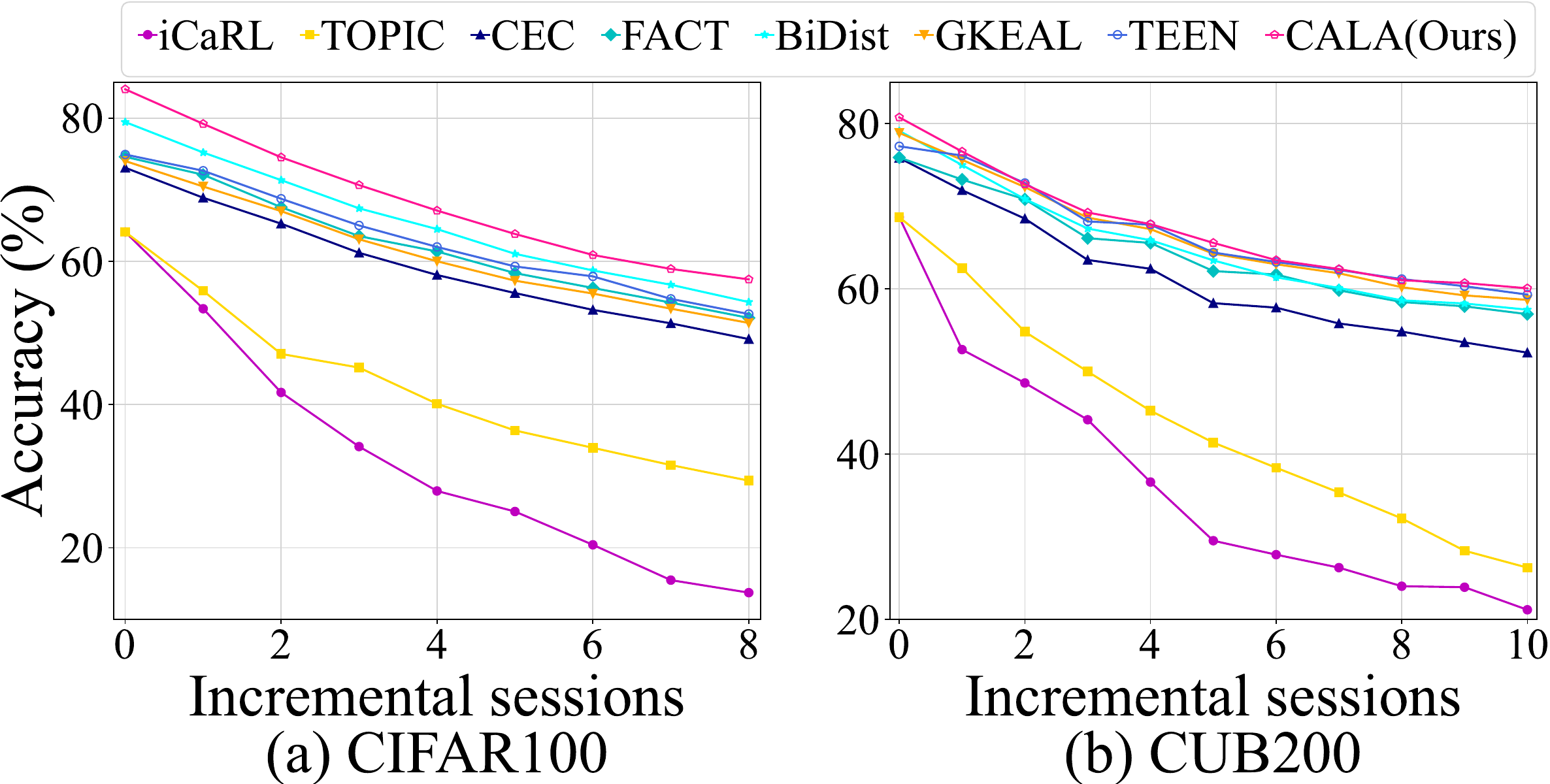} 
\caption{Comparison with the state-of-the-art works on the other two benchmarks: (a) CIFAR100 and (b) CUB200.}
\label{figcurve}
\end{figure}

Next, we evaluate our method's performance against several state-of-the-art works on the three benchmarks. Comprehensive data for \textit{mini}-ImageNet is shown in Table~\ref{tabminicom}. We select an optimal class-agnostic logit adjustment factor for LA-fscil to make a marginal improvement, which is defective. Furthermore, attributed to our class-aware logit adjustment strategy, CALA achieves the highest accuracy both in average and in the last session. The final results of CIFAR100 and CUB200 are presented in the form of performance curves in Fig.~\ref{figcurve}, showing CALA consistently outperforms the state-of-the-art results on average.

\subsection{Ablation Study}
\begin{figure}[ht]
\centering
\includegraphics[width=1\columnwidth]{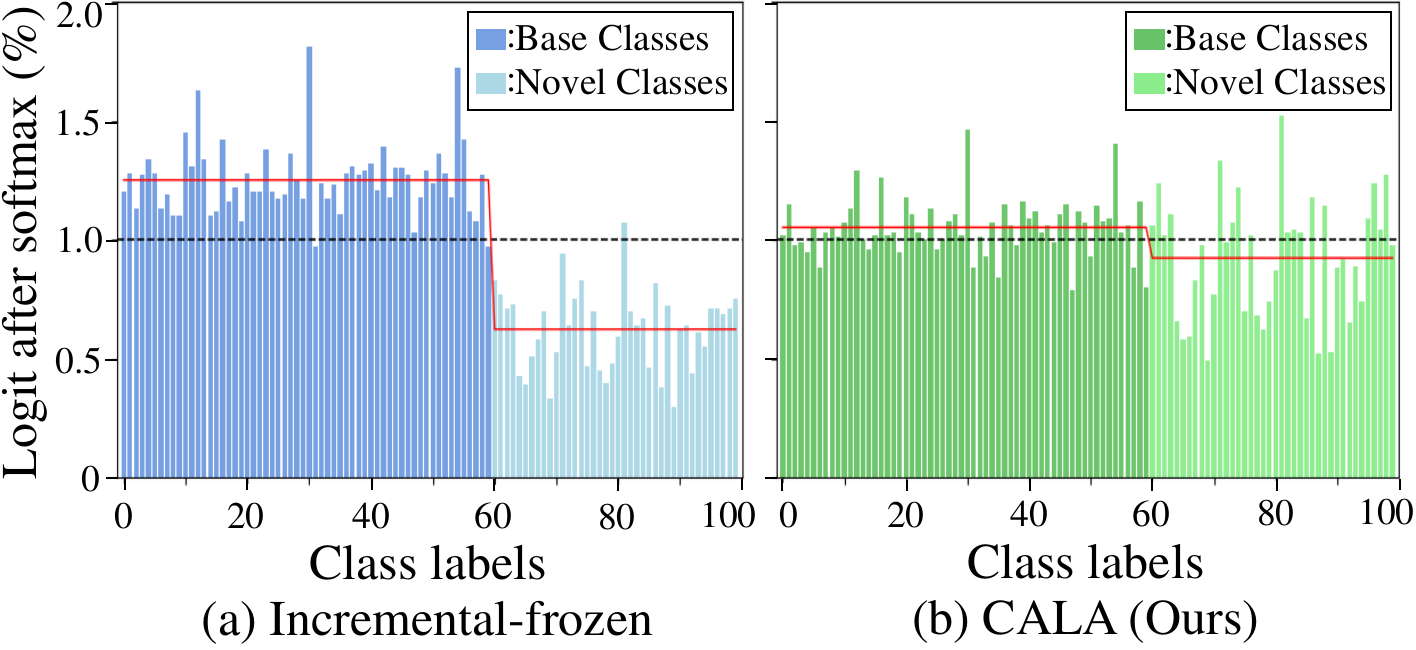} 
\caption{Ablation study on the logit distribution before and after our class-aware logit adjustment strategy on CIFAR100. (a) The logit distribution of a baseline is based on the incremental-frozen framework which lacks logit adjustment. (b) The logit distribution of the same baseline with the logit adjustment of our method.}
\label{figlogit}
\end{figure}

In the beginning, we conduct an ablation study on CIFAR100 to show the difference in logit distribution before and after logit adjustment. As shown in Fig.~\ref{figlogit}, we use the logit after a softmax function, \textit{i.e.}, $p_j = \frac{e^ {\mathbf{\hat{z}}_j}} {\sum_{k=1}^{N_C} e^{\mathbf{\hat{z}}_k}}$ as the measure of logit distribution. The black dashed line represents an extreme case where the predicted probabilities are the same across all classes. The red line parts above and below the black dashed line represent the mean logits for the base classes and novel classes, respectively. Obviously, in Fig.~\ref{figlogit} (a) the logit distribution of base classes overwhelms that of novel classes, which is caused by the traditional incremental-frozen framework, the over-learning of sufficient base knowledge leads to the classifier confusing novel instances into base classes. The imbalance of logit between base classes and novel classes is clear evident. However, after our CALA method, the imbalance gets suppressed. Intuitively speaking, the red lines better fit the black dashed line, and the weights of the excessively focused base classes are reduced, while those of the neglected novel classes are increased during classification. In consequence, CALA gets a better performance than the incremental-frozen baseline.

    
  
  

  

\begin{table}[h] 
\scriptsize
\centering
\renewcommand{\arraystretch}{0.7} 
\resizebox{0.7\linewidth}{!}{
 \begin{tabular}{l ccc}
  \toprule
  $\gamma$ & \multicolumn{1}{c}{$\text{Acc}_{\text{last}}$(\%)} & {Avg} & {$\Delta_{\text{last}}$} \\            
  \midrule
  $\gamma = 0$ & 55.42 & 67.11 & - \\
  $\gamma = 1$ & 55.59 & 67.16 & {+0.17} \\
  $\gamma = 10$ & \textbf{56.64} & \textbf{67.92} & \textbf{+1.22} \\
  $\gamma = 100$ & 26.53 & 42.31 & {-28.89} \\
  \bottomrule
\end{tabular}}
\caption{Ablation on the hyper-parameter $\gamma$ on CIFAR100. $\text{Acc}_{\text{last}}$ means the accuracy in the last session. $\Delta_{\text{last}}$ means the improvement in the last session.}
\label{tabcifargamma}	
\end{table}

\begin{table*}[t]
\scriptsize
\centering
\renewcommand{\arraystretch}{1.225} 
\resizebox{0.98\linewidth}{!}{
 \begin{tabular}{l ccccccccc cc}
  \toprule
  \multirow{2}{*}{Method} &\multicolumn{9}{c}{Accuracy in each session(\%)} & \multirow{2}{*}{$\text{HM}_{\textit{last}}$} & \multirow{2}{*}{$\text{Avg}_{\textit{novel}}$}\\
  \cmidrule{2-10}
  & {0} &{1}  &{2} & {3} &{4} &{5}  &{6} &{7} &{8}& & \\ 
  \hline              
  
  CEC{$^\diamond$}\cite{CEC}     & 72.17& 66.97 & 62.96 &59.50 &56.62 &53.78 &51.15 &49.32 &47.45 &27.80 &18.63  \\
  
  \textit{Improvement with CALA} (Ours) & - & \textbf{+0.06} & \textbf{+0.21} & \textbf{+0.35} & \textbf{+0.71} & \textbf{+0.81} & \textbf{+0.96} & \textbf{+1.07} & \textbf{+1.18} & \textbf{+9.60} & \textbf{+11.83}\\

  \midrule
    
  BiDist{$^\diamond$}~\cite{BiDist}  &74.60 &69.85 &65.30  &61.67 &58.65 &55.47 &52.77 &50.82 &49.01 &36.92 &28.41 \\
  
  \textit{Improvement with CALA} (Ours) & - & \textbf{+0.06} & \textbf{+0.23} & \textbf{+0.38} & \textbf{+0.75} & \textbf{+0.86} & \textbf{+1.02} & \textbf{+1.14} & \textbf{+1.36} & \textbf{+9.19} & \textbf{+13.68}\\
  
  \midrule

  SAVC{$^\diamond$}~\cite{savc}  &80.47  &75.55  &71.26  &67.66  &64.49  &60.50  &57.90  &55.75  &54.04  &45.40  &40.68\\
  
  \textit{Improvement with CALA} (Ours) & - & \textbf{+0.08} & \textbf{+0.15} & \textbf{+0.27} & \textbf{+0.49} & \textbf{+0.62} & \textbf{+0.75} & \textbf{+0.87} & \textbf{+1.02} & \textbf{+3.97} & \textbf{+7.72}\\

  \bottomrule
\end{tabular}}
\caption{The improvement of each session on three strong baselines without and with our plug-and-play module CALA. 
$\text{HM}_{\textit{last}}$ denotes the harmonic mean of the performances separately on base and novel classes in the last session. 
$\text{Avg}_{\textit{novel}}$ means the average performance only in novel classes. 
Results with $\diamond$ are reproduced based on their open-source codes.}

\label{tabminibeta}	
\end{table*}

Then we continue to use CIFAR100 and make a hyper-parameter analysis on $\gamma$, a hyper-parameter to limit $\mathbf{\beta}$ and avoid both the under-balancing and the over-balancing of logit. As shown in Table~\ref{tabcifargamma}, our method set $\gamma$ = 10, which can make a relatively best limitation on $\mathbf{\beta}$. In contrast, when $\gamma$ = 1, there is almost no difference from the baseline ($\gamma$ = 0), in this condition, $\gamma$ can't bring $\mathbf{\beta}$ to the proper magnitude, which leads to under-balancing. When $\gamma$ = 100, over-balancing occurs, although many novel samples confused as base samples are corrected in this condition, a large number of base samples are misclassified into novel classes instead, which causes a sharp decline in performance.

\subsection{Further Study}
\textbf{Plug-and-play Study.} We use CALA as a plug-and-play module on three strong baselines using the incremental-frozen framework on \textit{mini}-ImageNet, as shown in Table~\ref{tabminibeta}. We calculate a balance factor for each baseline, resulting in slight accuracy improvements in all of them. Notably, there are no gains in the base session since novel classes has not come, so no confusion arises. However, as incremental sessions proceed, the improvements become more noticeable due to more novel instances being correctly classified into their own classes. Although we did not specifically optimize hyper-parameters, we get a great improvement on the harmonic mean of base and novel performance in the last session, which presents CALA adjust all the base and novel logits into an equal level. Besides, we get obvious improvements in the average performance of novel classes across all sessions, which means CALA actually works as a plug-and-play module to adjust the misclassified novel instances.

\begin{figure}[h]
\centering
\includegraphics[width=1\columnwidth]{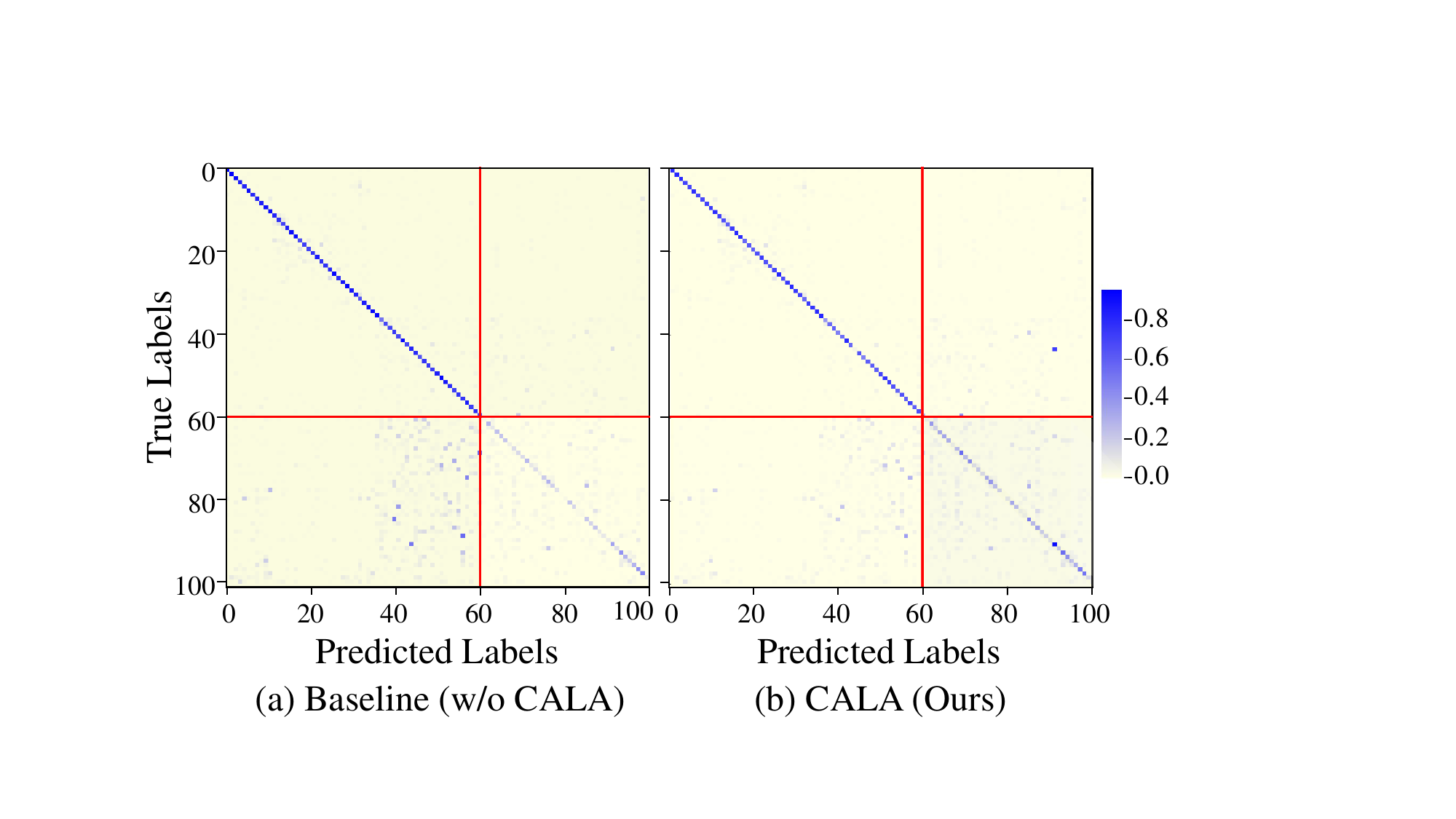} 
\caption{Confusion matrices without and
with CALA on \textit{mini}-ImageNet. Red lines are used to separate base classes and novel classes. CALA effectively improves the prediction accuracy in novel classes, resulting in a bluer diagonal in the last 40 classes.}
\label{figcm}
\end{figure}

\textbf{Confusion matrix.} To further observe the performance of CALA, we plot the confusion matrix without and with our class-aware logit adaptation module in Fig.~\ref{figcm}. The baseline without CALA generates a confusion matrix, which is scattered in the bottom left corner, meaning that the model tends to predict novel instances into base classes. After the adaption of CALA, the bottom left corner gets sparser, which represents there are fewer novel instances confused into base classes. Besides, the diagonal of novel classes gets bluer, meaning CALA actually reaches a higher performance on novel classes.

\begin{figure}[h]
\centering
\includegraphics[width=1\columnwidth]{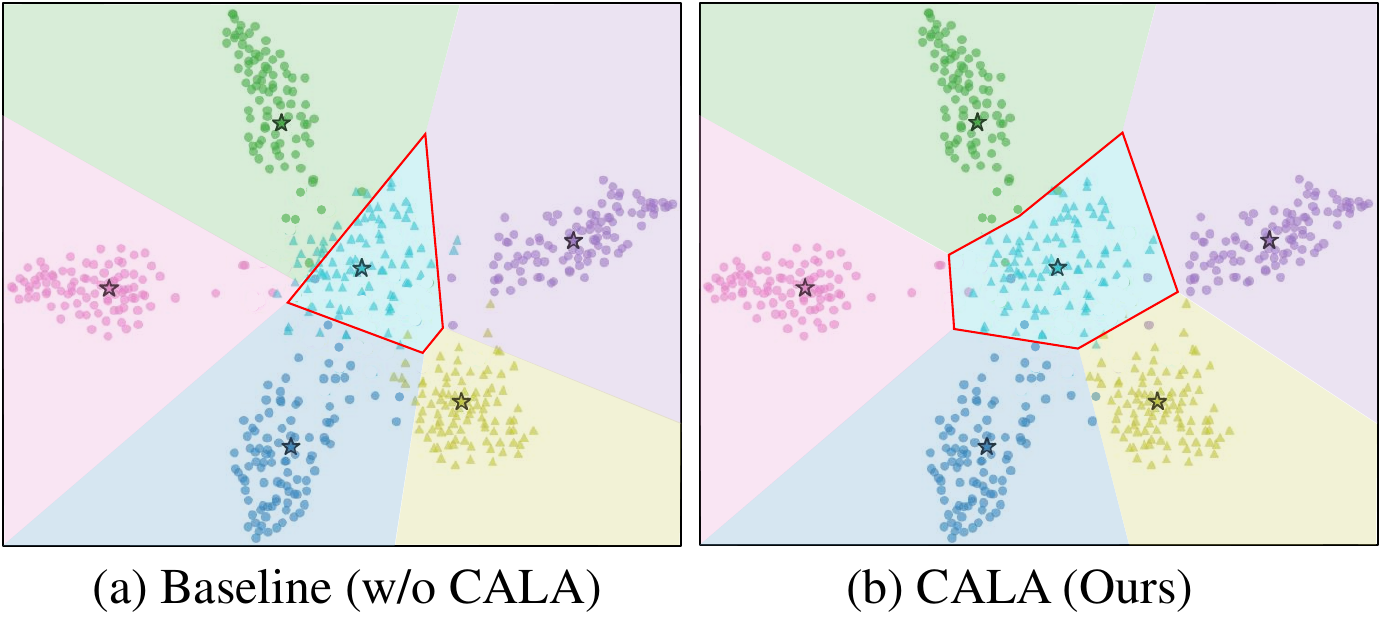} 
\caption{t-SNE \cite{tsne} visualization of feature space without and with CALA. Dots and triangles represent instances from base and novel classes separately. Different colors represent different classes. Stars indicate the centroids of classes. Geometric backgrounds with corresponding colors are decision boundaries. CALA enlarges the decision boundary from the novel class to the biased base class to include confused instances.}
\label{figtsne}
\end{figure}

\textbf{Visualization of adaption.} We plot the feature space and decision boundary in low-dimension space with t-SNE \cite{tsne} in Fig.~\ref{figtsne}. We randomly choose four base classes and two novel classes from CUB200. As shown in Fig.~\ref{figcm}, CALA works on the classifier level, thus making no difference to the feature space. The red boundary is the decision boundary of the novel class painted in light blue. Before adaption, class boundaries are biased to base classes, making instances of the novel class largely confused into base classes. After using CALA, the decision boundary of the novel class is enlarged to a more proper position, which adjusts the confused novel instances to their own classes.

\begin{figure}[h]
\centering
\includegraphics[width=1\columnwidth]{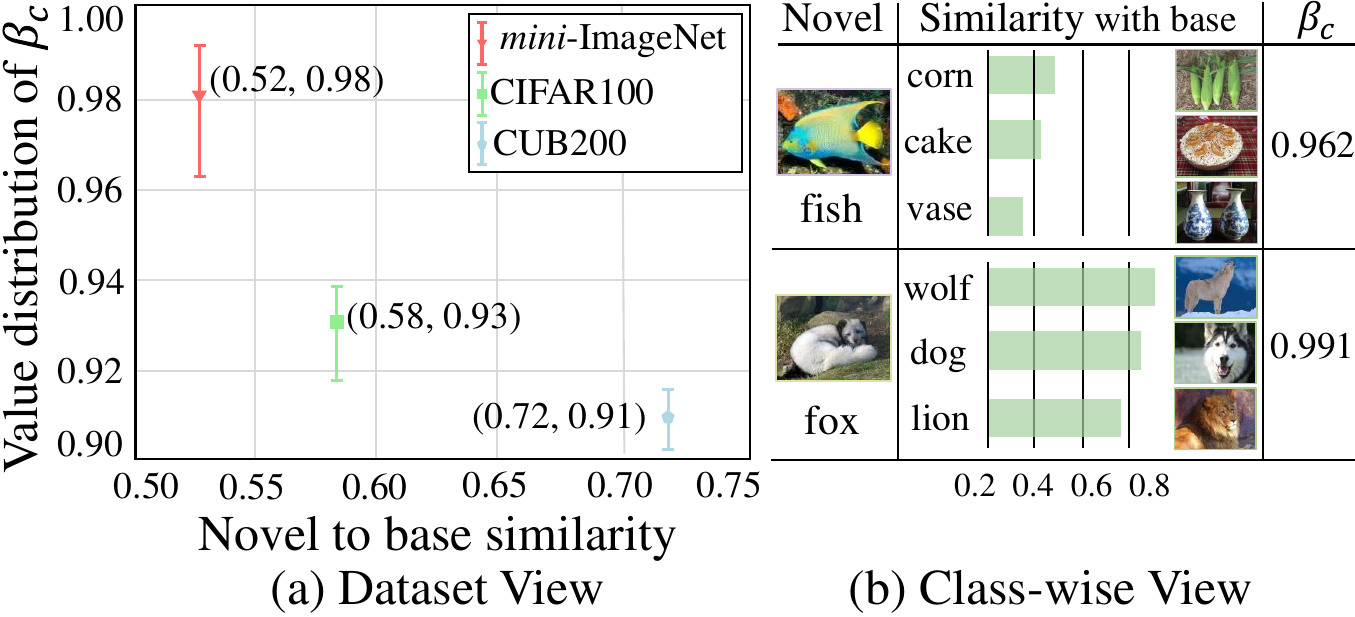} 
\caption{Further study on the relationship between similarity
and elements of $\mathbf{\beta}$: (a) analysis on all datasets (b) analysis on classes in $\textit{mini}$-ImageNet.}
\label{figsim}
\end{figure}

\textbf{Similarity analysis.} We further explore the relationship between novel-to-base similarity and each element $\beta_c$ calculated by the adapter. In Fig.~\ref{figsim} (a), error bars illustrate this relationship across three benchmark datasets. When novel classes are more similar with base classes, the mean value of $\beta_c$ increases, as we assign higher $\beta_c$ to novel classes that are more similar to base classes for stronger adjustment. Additionally, greater similarity is associated with a shorter error bar within the corresponding dataset, indicating a more stable distribution of $\mathbf{\beta}$ elements and a reduced need for adjustment. For example, CUB200, a fine-grained dataset, shows higher similarity, a lower mean $\beta_c$, and shorter error bars. We also conduct a class-wise analysis on $\textit{mini}$-ImageNet in Fig.~\ref{figsim} (b), where novel classes with higher overall similarity to base classes consistently have higher $\beta_c$, indicating a stronger requirement for adjustment.

\vspace{-0.5em}
\section{Conclusion}
In this paper, we propose a \textbf{C}lass-\textbf{A}ware \textbf{L}ogit \textbf{A}dapter (CALA), which realigns novel instances confused into base classes back to their own classes. CALA can work as a plug-and-play module and be integrated into any incremental-frozen framework. Additionally, different forms of experiments demonstrate the strong performance of CALA. \textbf{Limitations:} Our method occasionally misclassifies some base instances into novel classes. Although this is consistent with the FSCIL idea of trading base accuracy for the performance improvement of novel classes, we plan to optimize it from this perspective in our future works.

{
    \small
    \bibliographystyle{ieeenat_fullname}
    \bibliography{main}
}


\end{document}